\documentclass[final]{siamltex}
\usepackage{psfrag,graphicx,epsf}
\usepackage{amssymb}

\newtheorem{cor}[theorem]{Corollary}

\newtheorem{example}[theorem]{Example}
\newtheorem{alg}[theorem]{Algorithm}
\newtheorem{sub}[theorem]{Subroutine}

\newcommand{\df}[1]{{\bf{#1}}}

\newcommand{\real}{\mathbb{R}}

\newcommand{\qed}{$\diamond$}

\newcommand{\eg}{{\em e.g.}}
\newcommand{\ie}{{\em i.e.}}

\newcommand{\pf}{{\sc proof: }}
\newcommand{\inv}{{^{-1}}}
\newcommand{\rest}[2]{{\left.{#1}\right\vert_{#2}}}
\newcommand{\del}{\partial}
\newcommand{\CC}{{\mathcal C}}

\newcommand{\graph}{\Gamma}

\renewcommand{\L}{{\mathcal L}}
\newcommand{\catalogue}{{\sf C}}
\newcommand{\workspace}{{\mathcal W}}

\newcommand{\scpx}{{\mathcal S}}
\newcommand{\shcpx}{{\mathcal Sh}}
\newcommand{\obstacle}{{\mathcal O}}
\newcommand{\state}{{U}}
\newcommand{\locstate}{{\hat{U}}}
\newcommand{\gen}{{\mathbf{\phi}}}
\newcommand{\action}{{\mathbf{\Phi}}}
\newcommand{\link}[1]{\ell k[{#1}]}
\newcommand{\trace}{{\mbox{\sc tr}}}
\newcommand{\support}{{\mbox{\sc sup}}}

\newcommand{\St}{{\sf St}}

\begin{document}

\pagestyle{myheadings}
\thispagestyle{plain}
\markboth{AARON ABRAMS \& ROBERT GHRIST}{COMPLEXES FOR METAMORPHIC ROBOTS}

\title{State Complexes for Metamorphic Robots\footnote{
AA supported by National Science Foundation Grant DMS-0089927.
RG supported by National Science Foundation Grant DMS-0134408.}}
\author{A. Abrams\thanks{
        Department of Mathematics, University of Georgia, 
        Athens, GA 30602, USA} 
\and
        R. Ghrist\thanks{
	  Department of Mathematics, University of Illinois, 
	  Urbana, IL 61801, USA}}
\maketitle

\begin{abstract}
A \df{metamorphic} robotic system is an aggregate of homogeneous robot 
units which can individually and selectively locomote
in such a way as to change the global shape of the system. We introduce a 
mathematical framework for defining and analyzing general metamorphic
robots. This formal structure, combined with ideas from geometric
group theory, leads to a natural extension of a configuration space
for metamorphic robots --- the \df{state complex} --- 
which is especially adapted to parallelization.
We present an algorithm for optimizing reconfiguration sequences
with respect to elapsed time. A universal geometric property of 
state complexes --- \df{non-positive curvature} --- is the key to 
proving convergence to the globally time-optimal solution. 
\end{abstract}

\section{Introduction}

\begin{figure}[hbt]
\begin{center}
\includegraphics[angle=0,width=3.5in]{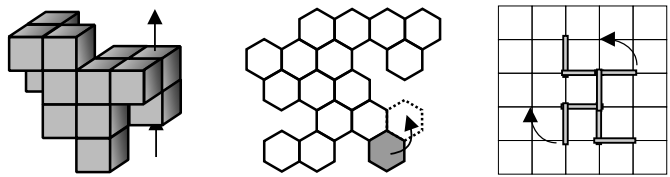}
\caption{Metamorphic systems may be built on a variety of lattice
structures with sliding or pivoting motion.}

\label{fig_Reconfig}
\end{center}
\end{figure}

In recent years, several groups in the robotics community have been
modeling and building \df{reconfigurable} or, more specifically, 
\df{metamorphic robots} (\eg, \cite{Chi1,KR,Murata1,Yim1,Yim2}).
Such a system consists of multiple identical robotic
cells in an underlying lattice structure
which can disconnect/reconnect with adjacent neighbors, and slide,
pivot, or otherwise locomote to neighboring lattice points following
prescribed rules: see Fig.~\ref{fig_Reconfig}. There are as many
models for such robots as there are researchers in the sub-field: 
2-d and 3-d lattices; hexagonal, square, and dodecahedral cells;
pivoting or sliding motion: see, \eg,  
\cite{Chi1,KR,MR,Murata1,Murata2,BKRT,Guibas,Yim1,Yim2} and the references
therein. The common feature of these robots is
an aggregate of lattice-based cells having prescribed local 
transitions from one shape to another.

The primary challenge for such systems is \df{shape-planning}: 
how to move from one shape to another via legal moves. One
centralized approach \cite{Chi2,PEC} is to build a 
\df{transition graph} whose vertices are the various shapes 
and whose edges are elementary legal moves from
one shape to the next. It is easily demonstrated that the size of 
this graph is exponential in the number of cells. 

We propose to extend the notion of a configuration space to metamorphic
robots in a novel manner. The idea: consider the transition 
graph described above as a one-dimensional skeleton of a higher-dimensional
cubical complex, the \df{state complex}. Assume that from a given state 
there are two legal moves which are physically 
independent (or, more suggestively, ``commutative''): \ie, these moves can be
executed simultaneously. In the transition graph, this corresponds to
the four edges of a square. For any pair of commutative moves, fill in 
the four edges of the graph with an abstract square
2-cell. Continue inductively adding $k$-dimensional cubes corresponding 
to $k$-tuples of physically independent motions. The result is a cubical 
complex which has several advantages over the transition graph:
\begin{enumerate}
\item 
{\bf Simplicity.}  
  The state complex is often simpler than the
  transition graph: \eg, the 1-d graph of an $n$-dimensional cube has 
  $n2^{n-1}$ edges. This figure belies the simplicity of the single cube. 

\item
{\bf Speed.}
  Geodesics on this complex cut across the diagonals of cubes whenever 
  possible. One performs all possible commutative motions
  simultaneously, maximizing parallelization and yielding a speed-up 
  by a factor equal to the number of coordinated motions.

\item
{\bf Shape.}
  The global geometry/topology of the state complex 
  carries information about the metamorphic system. 
  For certain examples, the topology of the state complex 
  ``converges'' upon refining the lattice. In addition, only 
  special geometries can be realized as the state complex of 
  a local metamorphic system: commutativity in reconfiguration leads 
  to an abhorrence of positive curvature in the state complex.
\end{enumerate}

Sections~\ref{sec_Def} through \ref{sec_Shape} give definitions and 
examples of [abstract] metamorphic systems and their state complexes. The 
next two sections (Sections~\ref{sec_Topology}-\ref{sec_Geometry})
detail topological and geometric features of the state complex. 

For large systems, the problem of computing the state complex and 
designing geodesics in order to perform shape planning is computationally
infeasible, primarily because the size of the
complex is often exponential in the number of robot cells. In addition,
any control scheme induced by geodesic construction is necessarily
centralized. Several researchers have begun building 
decentralized control algorithms for shape planning 
\cite{BBR,Walter2,Walter1,Yim3,Yoshida}. 
Such algorithms have the advantage of speed and scalability; 
however, the reconfiguration paths are typically not optimal.

As an application of our techniques, we present 
in Sections~\ref{sec_Time}-\ref{sec_Alg} 
an algorithm for trajectory optimization which takes as its argument 
an arbitrary edge path in the transition graph. Algorithm~\ref{alg_Short} 
then performs a type of \df{curve shortening} within the state
complex. A deep theorem about the curvature of all state complexes
(Theorem~\ref{thm_NPC}) is then used to prove that this
algorithm returns a shape trajectory which is the global minimum
obtainable from this path with respect to elapsed time. 

Our definitions and theorems are phrased for systems involving 
``discrete'' reconfiguration. More general types of robots
which employ continuous reconfiguration for locomotive gaits 
(such as the Polybot developed by M. Yim's lab at Xerox PARC) 
are not covered by our definitions. We note, however, that 
certain locomotive reconfigurable robots can be thought of as
lattice-based tiles by amalgamating subsystems \cite{MR}. In 
addition, our definitions can easily be extended to more general
non-lattice reconfigurable systems \cite{AG:state}. 

\section{A mathematical definition}
\label{sec_Def}

While it is easy to generate examples of what is meant by a metamorphic
robot, it is more challenging to write a clean mathematical definition. 
We propose a set of definitions which is broad enough to include 
some non-obvious examples. The paper \cite{BKRT} suggests a similar
type of structure using cellular automata rule sets. 

A local metamorphic system is a collection of  
states on a lattice, where each state is thought of as an indicator 
function for the aggregate. Any state can be modified by local 
rearrangements, these local changes being coordinated by a
catalogue of models realized under the actions of isometries into
the workspace. The adjective ``local'' refers to legality criteria: 
anywhere in the workspace at which a local change from the catalogue 
can be applied, it is legal to do so. To incorporate obstacles and 
basepoints into our systems, we distinguish between the amount of 
information needed to determine the legality of an elementary move 
(the ``support'' of the move) and the precise place in which modules 
are actually in motion (the ``trace'' of the move). 

\begin{definition}{
\label{def_generator}
Let $\L$ denote a lattice in $\real^k$ and let $\workspace\subset\L$ be
some workspace. The \df{catalogue} $\catalogue$ for a local metamorphic 
system on $\workspace$ is a collection of \df{generators}. 
Each generator $\gen\in\catalogue$ consists of (1) the \df{support}, 
$\support(\gen)\subset\L$; (2) the \df{trace} of the 
move, $\trace(\gen)\subset\support(\gen)$; and (3) an unordered
pair of local states $\locstate_{0,1}:\support(\gen)\to\{0,1\}$ 
satisfying\footnote{
All generators are assumed to be \df{nondegenerate} in the sense
that $\locstate_0\neq\locstate_1$.}
\begin{equation}
\label{eq_locstates}
       \rest{\locstate_0}{\support(\gen)-\trace(\gen)} 
   =   \rest{\locstate_1}{\support(\gen)-\trace(\gen)}.
\end{equation}
Otherwise said, the local states are equal on $\support(\gen)-\trace(\gen)$. 
}\end{definition}

\begin{definition}{
\label{def_action}
An \df{action} of a generator $\gen\in\catalogue$
is a rigid translation $\action:\support(\gen)\hookrightarrow\workspace$. 
Given a state $\state:\workspace\to\{0,1\}$, an action $\action$ 
is said to be \df{admissible} at $\state$ if $\locstate_0=\state\circ\action$. 
In this case, we write 
\[
        \gen[\state] := \left\{
        \begin{array}{cl}
           \state & : {\mbox{ on }} \workspace-\action(\support(\gen))\\
           \locstate_1\circ\action\inv & : {\mbox{ on }} \action(\support(\gen)) 
        \end{array} \right. .
\]
(Note that $\action$ is left out of the notation.)
}\end{definition}

\begin{definition}{
A \df{local metamorphic system} on $\workspace$ is a collection of finite
states $\{\state_\alpha:\workspace\to\{0,1\}\}$ closed under all possible 
admissible actions of generators in the catalogue $\catalogue$. 
(A state is finite if $\state=1$ at only finitely many points of
$\workspace$.)
}\end{definition}

To repeat, the catalogue and the workspace are the ``seeds'' for
a local metamorphic system. From this pair, all possible translations of 
the supports into $\workspace$ yield the actions. Then, a 
collection of states on the workspace is a local metamorphic
system if, whenever an action $\action$ of a generator $\gen$ 
on a state $\state$ is admissible, then the corresponding
state $\gen[\state]$ is also included. 

A metamorphic system with \df{obstacles} $\obstacle\subset\workspace$ 
satisfies in addition 
\begin{equation}
        \action(\trace(\gen))\cap\obstacle=\emptyset , 
\end{equation}
for each $\gen\in\catalogue$.
Obstacle sets count as legal positions for determining the admissibility
of a move (the support of an action may intersect $\obstacle$), but
no motion of metamorphic agents may incorporate the obstacle sites
(the trace of an action must not intersect $\obstacle$).

\section{Examples}
\label{sec_Ex}
Some of the following examples are inspired by metamorphic robots
already developed; other examples are more abstract. 

\begin{figure}[hbt]
\begin{center}
\includegraphics[angle=0,width=4.5in]{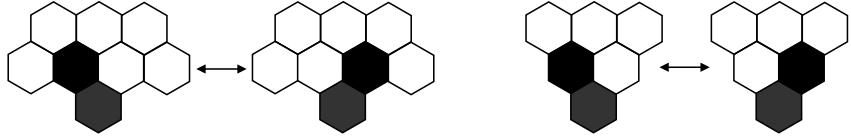}
\caption{Two different catalogues for a 2-d hexagonal lattice system 
with pivots. Black cells are occupied, white are unoccupied. 
The local states $\locstate_0$ and $\locstate_1$ are shown for 
each generator.}

\label{fig_HexEx}
\end{center}
\end{figure}

\begin{example}{\em
\label{ex_Hex}[2-d hex with pivots]
We present two slightly different catalogues, each with six generators
(or one, up to discrete rotations), in Fig.~\ref{fig_HexEx}. Both
of these systems, modeled after that of \cite{Chi1}, have local 
moves which pivot a planar hexagon about a neighbor. For all generators 
presented, the trace is equal to the two central hexagons. In the 
first system, the support is chosen so that the aggregate does not change its
topology but only its shape. The slightly smaller support of the second
catalogue allows for local topology changes. To model
a fixed ``base'' cell (which is, say, affixed to a power source
as in \cite{Guibas}), one establishes this cell as an obstacle 
$\obstacle$. 
}\end{example}

\begin{example}{\em
\label{ex_SquareSlide}[2-d square lattice]
In Fig.~\ref{fig_SquareEx}, we display a generator for a 
planar system in which rows [as pictured] and columns [not pictured] 
of an aggregate of square cells can slide. There
are in fact several generators represented in ``shorthand,'' one
for each $k\geq 0$. A dot inside a cell indicates that it can
be either occupied or unoccupied, but if occupied, then its 
neighbor (indicated by an arrow) must also be turned on. This condition 
guarantees that the aggregate does not disconnect (even locally)
under slides. The trace of this set of generators is the 
entire middle row except the two endpoints. To keep 
the catalogue finite, one would include only those 
generators with $k\leq N$, where $N$ is the number of 
occupied cells in any state.

\begin{figure}[hbt]
\begin{center}
\psfragscanon
\psfrag{k}[bc][bc]{\Large{$k$}}
\psfrag{0}[c][c]{\Large{$0$}}
\psfrag{1}[c][c]{\Large{$1$}}
\includegraphics[angle=0,width=4.9in]{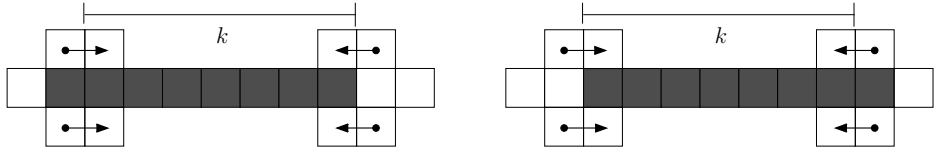}
\caption{The row-sliding generators for a sliding-squares system.}

\label{fig_SquareEx}
\end{center}
\end{figure}

}\end{example}

\begin{example}{\em[2-d articulated planar arm]
\label{ex_RobotArm}
Consider as a workspace $\workspace$ the set of edges in the planar
integer lattice. The catalogue consists of two generators,
pictured in Fig.~\ref{fig_ArmEx}. Beginning with a state
having $N$ vertical edges end-to-end, 
the metamorphic system thus generated
models the position of an articulated robotic arm with fixed base which can 
(1) rotate at the top end and (2) flip corners as per the diagram.

\begin{figure}[hbt]
\begin{center}
\includegraphics[angle=0,width=5.3in]{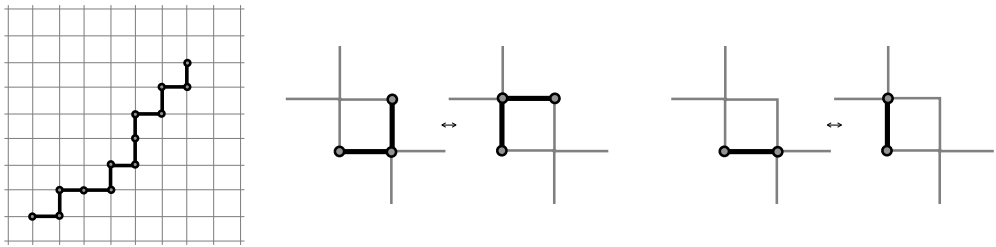}
\caption{A positive articulated robot arm [left]. 
One generator [center] flips corners and has as its 
trace the central four edges. The other generator [right] rotates 
the end of the arm, and has trace equal to the two activated edges.}

\label{fig_ArmEx}
\end{center}
\end{figure}
If one includes rotations of these generators, more intricate types
of configurations are possible, including deadlocked configurations. 
Highly self-reconfigurable examples with multiple interacting arms 
can be realized by adding new generators: an ``attach-detach'' 
generator which allows endpoints of arms to merge (thus yielding 
a ``marked point'' at the attachment); and a ``sliding''
generator which allows this marked point to slide, having the effect
of allowing the attached arms to trade segments. These 
generators are pictured (up to Euclidean symmetries) in 
Figure~\ref{fig_ArmEx2}.

\begin{figure}[hbt]
\begin{center}
\includegraphics[angle=0,width=5.2in]{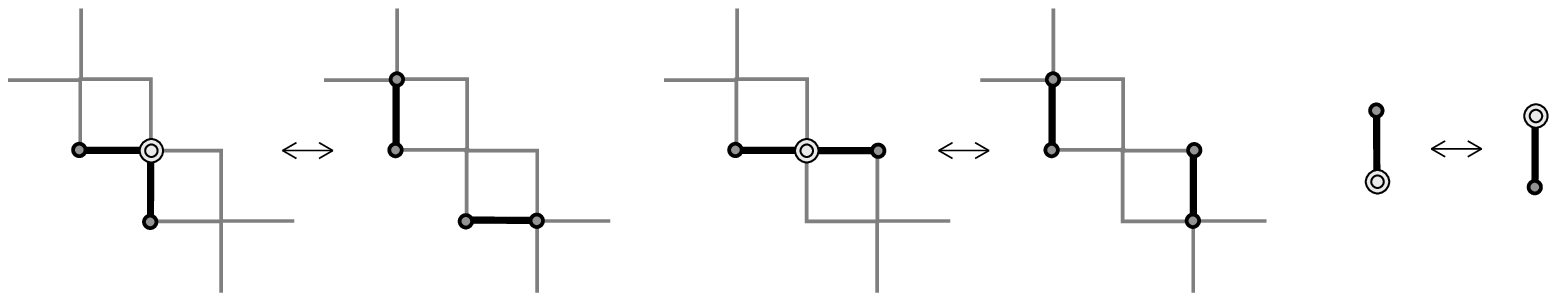}
\caption{Additional generators allow for attachment and 
detachment of arm endpoints [left, center] and sliding of the coupling 
points [right].}

\label{fig_ArmEx2}
\end{center}
\end{figure}

}\end{example}


One of the benefits of writing down a rigorous definition of a metamorphic
system is the discovery of systems which have little resemblance to 
the systems of, say, Fig.~\ref{fig_Reconfig}.  In particular, our
definitions easily extend to metamorphic systems which are not 
lattice-based; the following example is especially interesting.

\begin{example}{\em
\label{ex_DSpaceGraph}
Consider a finite graph $\Gamma$ in which every edge is assigned a length
of one.  (Every graph can be embedded in some $\real^k$ so as to have
this property.)  The catalogue consists
of a single generator whose support and trace are precisely the closure 
of a single ``abstract edge.''  The local states of this generator 
consist of the pair $\locstate_0$ and 
$\locstate_1$ which evaluate to $1$ on one of the endpoints and $0$ 
on the other.  The actions in this case are length-preserving maps
from the abstract edge into $\Gamma$.
The metamorphic system generated from a state
$\state$ on $\Gamma$ with $N$ vertices evaluating to $1$ mimics an
ensemble of $N$ unlabeled non-colliding Automated Guided Vehicles on 
$\Gamma$, cf. \cite{GK}. 
}\end{example}

More abstract examples of metamorphic examples include spaces of
triangulations of polygons with edge-flipping as the generator,
examples arising from word representations in group theory, and certain
multi-step assembly processes \cite{AG:state}.

\section{The state complex}
\label{sec_Shape}

In the robotics literature, one often models a configuration space
for a metamorphic system with a transition graph which represents
actions of elementary moves on states. That is, the vertex set is the 
collection of all states $\{\state_\alpha\}$, and the edges are 
unoriented pairs of states which differ by the action of one generator. 
Transition graphs are discussed for shape-planning in several 
particular cases in the literature (planar hex case: \cite{Chi2,PEC}).  
Our departure is to make the transition graph the 1-skeleton of 
a cubical complex (an analogue of a simplicial complex, but made out
of abstract cubes) which coordinates parallel or ``commutative'' motions. 

\begin{definition}{
\label{def_Commute}
In a local metamorphic system, a collection of actions of 
(not necessarily distinct) generators $\{(\gen_{\alpha_i},
\action_{\alpha_i})\}$ is said to \df{commute} if
\begin{equation}
\label{eq_Commute}      
        \action_{\alpha_i}(\trace(\gen_{\alpha_i}))\cap
        \action_{\alpha_j}(\support(\gen_{\alpha_j})) = \emptyset 
        \quad \forall i\neq j .
\end{equation}
}\end{definition}

\begin{example}{\em
Two simple examples suffice to illustrate the difference between 
commuting and noncommuting actions. First, consider the pair
of commuting moves for a planar hexagonal pivoting system, as 
represented in Fig.~\ref{fig_Commute} [left].
%
\begin{figure}[hbt]
\begin{center}
\includegraphics[angle=0,width=4.2in]{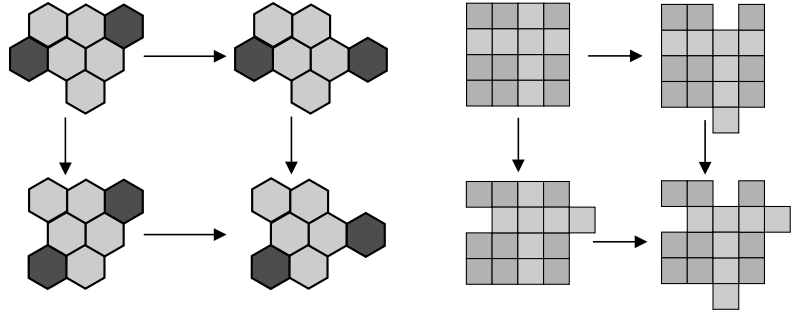}
\caption{Examples of commuting [left] and noncommuting [right] actions in 
planar systems.}

\label{fig_Commute}
\end{center}
\end{figure}
%
Compare this with a planar sliding block example as illustrated
in Fig.~\ref{fig_Commute} [right]. Although the pair of moves illustrated
forms a square\footnote{The individual robotic cells are not labeled:
only the shape of the aggregate is recorded.} 
in the transition graph, this particular pair 
of actions does not commute. Physically, it is obvious why these
moves are not independent: sliding the column part-way obstructs
sliding a transverse row. Mathematically, this is captured by
the traces of the actions intersecting.  
}\end{example}

The state complex has an abstract $k$-cube for each collection of
$k$ admissible commuting actions:

\begin{definition}{
\label{def_StateCpx}
The \df{state complex} $\scpx$ of a local metamorphic system
is the following abstract cubical complex.
Each abstract $k$-cube $e^{(k)}$ of $\scpx$ is an equivalence class
$[\state;(\action_{\alpha_i})_{i=1}^k]$ where
\begin{enumerate}
\item 
        $(\action_{\alpha_i})_{i=1}^k$ is a
        $k$-tuple of commuting actions of generators $\phi_{\alpha_i}$;
\item
        $\state$ is some state for which all the actions
        $(\action_{\alpha_i})_{i=1}^k$ are admissible; and
\item
$         [\state_0;(\action_{\alpha_i})_{i=1}^k]
        =[\state_1;(\action_{\beta_i})_{i=1}^k]$   
        if and only if the list $(\beta_i)$ is a permutation of $(\alpha_i)$
        and $\state_0=\state_1$ on the set
$        \workspace - \bigcup_i\action_{\alpha_i}(\support(\gen_{\alpha_i}))$ .
\end{enumerate}

The boundary of each abstract $k$-cube is the collection of 
$2k$ faces obtained by deleting the $i^{th}$ action from the list and 
using $\state$ and $\gen_{\alpha_i}[\state]$ as the ambient states.
Specifically,
\begin{equation}
        \del[\state;(\action_{\alpha_i})_{i=1}^k] = 
        \bigcup_{i=1}^k\left( 
        [\state;(\action_{\alpha_j})_{j\neq i}]
	\cup
        [\gen_{\alpha_i}[\state];(\action_{\alpha_j})_{j\neq i}]
	\right).
\end{equation}
}\end{definition}

It follows easily that the $k$-cells are well-defined with respect to 
admissibility of actions. 
The proof of the following obvious lemma is given in detail to
flesh out the previous definition.

\begin{lemma}
\label{lem_stupid}
(a) The 0-dimensional skeleton of $\scpx$, $\scpx^{(0)}$, is the 
set of states in the reconfigurable system.
(b) The 1-dimensional skeleton of $\scpx$, $\scpx^{(1)}$, is precisely 
the transition graph.
\end{lemma}

\pf~
(a) Vertices of $\scpx$ consist of equivalence classes
consisting of zero (i.e., no) actions of generators up to permutation, 
together with a state defined on the complement of the supports
of the actions. As there are no actions, each 0-cell is 
precisely a single state of the reconfigurable system.

(b) A 1-cell of $\scpx$ is an equivalence class of the 
form $[\state;(\action)]$. The only other representative
of the equivalence class is $[\gen[\state];(\action)]$; hence, 
the 1-cells are precisely the edges in the transition graph.
Clearly, the boundary of $[\state;(\action)]$ is the pair
of 0-cells $[\state; (\cdot)]$ and $[\gen[\state]; (\cdot)]$. 
\qed\vspace{0.1in}

For small numbers of cells, it is easy to illustrate the state complex.

%
\begin{figure}[hbt]
\begin{center}
\psfragscanon
\psfrag{p}[t][]{{\Huge $p$ }}
\psfrag{q}[l][]{{\Huge $q$ }}
\includegraphics[angle=0,width=4.0in]{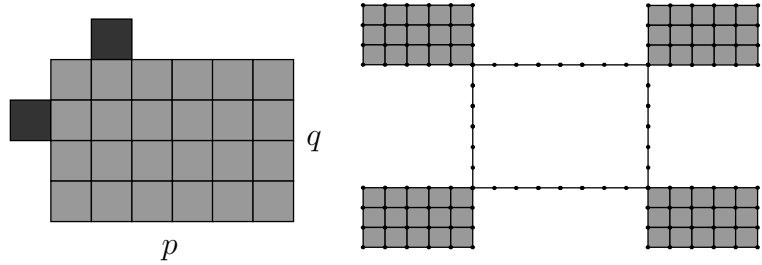}
\caption{The planar sliding square example [left] with two movable 
blocks [in black] and a $p$-by-$q$ obstacle set [in grey] yields 
a state complex $\scpx$ that is topologically a circle [right].} 

\label{fig_SqShEx}
\end{center}
\end{figure}
%
\begin{example}{\em
\label{ex_SCSlide}
Consider the 2-d square lattice row/column sliding system whose catalogue
is illustrated in Fig.~\ref{fig_SquareEx}. If we consider
a system with (occupied) obstacles in the form of a $p$-by-$q$ rectangle 
generated from the state of Fig.~\ref{fig_SqShEx} [left],
one obtains a planar transition graph with $4(pq+1)+2(p+q)$ vertices
and $8(pq+1)-2(p+q)$ edges. In contrast, the state complex is that
of Fig.~\ref{fig_SqShEx} [right]: this is topologically a circle, 
corresponding to the fact that the pair of free squares 
can circulate about the obstacle set through a sequence of slides. 
The large 2-d regions correspond to states in which the two free
squares are on separate (but adjacent) sides of the obstacle set.
}\end{example}

\begin{example}{\em
\label{ex_SCHex}
Consider the planar hex system of Fig.~\ref{ex_Hex} [left]
with a workspace $\workspace$ consisting of a long channel of four 
rows, the bottom row being filled obstacles, and the top row being 
non-occupied obstacles. A line of cells in one of the two free rows 
can ``climb across'' one by one, yielding the state complex of 
Fig.~\ref{fig_ShHexEx}[right] (cf. the algorithm of \cite{Walter1}). 
This state complex is contractible for any length column.
\begin{figure}[hbt]
\begin{center}
\includegraphics[angle=0,width=4.3in]{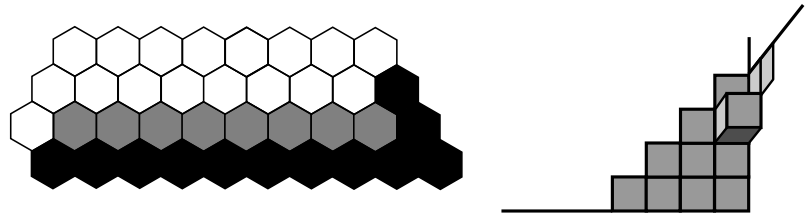}
\caption{For a line of hexagons filing out of a constrained tunnel 
[left], the state complex is contractible [right]. Black cells represent
occupied obstacle sets; the top row is an unoccupied obstacle set.
A fairly short tunnel is shown: for longer tunnels, the state complex
is still contractible, but of higher dimension.}

\label{fig_ShHexEx}
\end{center}
\end{figure}
}\end{example}

\begin{figure}[hbt]
\begin{center}
\includegraphics[angle=0,width=3.2in]{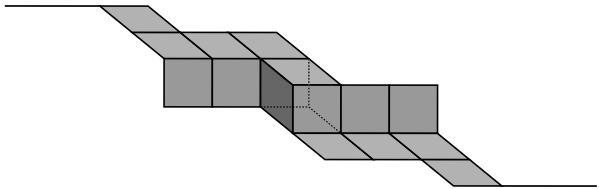}
\caption{The state complex of a 5-link positive arm has one 
cell of dimension three, along with several cells of lower
dimension.}

\label{fig_ShArmEx}
\end{center}
\end{figure}

\begin{example}{\em
\label{ex_SCRobotArm}
The state complex associated to the positive articulated robot arm 
of Example~\ref{ex_RobotArm} in the case $N=5$ is given in 
Fig.~\ref{fig_ShArmEx}. Note that there can be at most three
independent motions (when the arm is in a ``staircase''
configuration); hence the state complex has top dimension three.
Notice also that although the transition graph for this system
is complicated, the state complex itself is topologically trivial
(contractible). 
}\end{example}

\begin{example}{\em 
\label{ex_SCGraph}
In the system of Example~\ref{ex_DSpaceGraph} with the graph being 
a $K_5$ (the complete graph on five vertices) and $N=2$, the 
state complex is a two-dimensional closed surface. A
simple combinatorial argument (as in \cite{Abr00,AG:02})
reveals that the Euler characteristic
is $-5$, implying that the state complex is non-orientable.
If the AGV's are labeled, the state complex becomes a closed
orientable surface of genus 6.
}\end{example}

\section{The topology of $\scpx$}
\label{sec_Topology}

If one looks at a transition graph without knowing the particulars
of the metamorphic system, very little information can be 
extracted. This paper argues that completing the transition 
graph to the state complex is ``natural'' --- the state 
complex simplifies the transition graph and endows it with 
topological and geometric content.

Our first example of naturality is motivated by the desire to 
build metamorphic systems with large numbers of micro- or nano-scale
cells. While large numbers of cells would yield a type of 
continuum-limit convergence on the dynamics of shape change, 
the resulting transition graphs have no such convergence. The size of 
the transition graph goes up exponentially in the number of cells; 
more ominous is that the topology of the transition graph 
(the number of basic cycles) blows up as well. This is not always so with 
the state complex: in certain key examples, the topology of 
$\scpx$ is either invariant or converges to a limiting type. 

We have already seen one such example of this stabilization. In 
Example~\ref{ex_SCSlide}, the state complex of a pair of squares
sliding along a rectangular obstacle of size $p$-by-$q$ is 
topologically a circle, independent of $p$ and $q$. This can 
be interpreted as a type of convergence: consider the effect
of refining the underlying lattice structure, increasing $p$ 
and $q$ while maintaining a pair of sliding squares. Then the 
dimension of the state complex remains the same, as does the 
topological type of the space. Intuitively speaking, the 
``limit'' as this refining process is repeated yields a 
``topological'' configuration space of two points sliding 
smoothly along the boundary of a rectangle, which can dock or
un-dock at the corners. 

\begin{example}{\em
Recall the state complex associated to the metamorphic system of $N$
points on a graph $\graph$, Example~\ref{ex_DSpaceGraph}. Consider a 
refinement of $\graph$ which inserts additional vertices along edges. 
It follows from the techniques of \cite{Abr00}
that the state complex of this refined system has the same 
topological type (up to homotopy equivalence) after a fixed 
bound on the refinement ($N$ additional vertices per 
edge). Furthermore, this ``stabilized'' state complex is 
in fact homotopic to the topological configuration space 
of $N$ non-colliding points of $\graph$ --- precisely what one 
expects as the number of refinements goes to infinity. 
}\end{example}

\begin{example}{\em
Recall the positive articulated robot arm of Example~\ref{ex_RobotArm}. 
Consider a refinement of the underlying lattice which shrinks the 
lattice by a factor of two (or, equivalently, which inserts an 
additional joint in the middle of each edge). This is a more 
dramatic change since the dimension of the state complex doubles.
Nevertheless, the topological type is invariant:  the state
complex remains contractible.
\begin{proposition}
Let $\scpx_N$ denote the state complex of the positive articulated arm
from Example~\ref{ex_RobotArm} with $N$ segments. The complexes
$\scpx_N$ are all contractible. 
\end{proposition}

\pf~
For these articulated arms, there is a nice inductive structure
on the state complexes. Fixing $N$, each state (vertex) in $\scpx_N$ 
is represented as a length $N$ word in the symbols $x$ and $y$,
where $x$ denotes the arm going to the right and $y$ denotes the arm
going up. In this language, the two generators are (1) transposing 
a subword $xy\leftrightarrow yx$, and (2) changing the last letter of 
the word. 

Consider the subcomplex $X\subset\scpx_N$ consisting of all cells whose 
vertices have words beginning with the letter $x$. Likewise, let $Y$
denote the subcomplex all of whose vertices begin with the letter $y$. 
These subcomplexes are each a copy of $\scpx_{N-1}$ which we may 
assume inductively is contractible. One passes between the subcomplexes 
$X$ and $Y$ only when a move exchanges the initial two letters of
the word from $xy$ to $yx$. The connecting set is thus homeomorphic
to $\scpx_{N-2}\times[-1,1]$ and attached to $X$ and $Y$ along 
$\scpx_{N-2}\times\{-1\}$ and $\scpx_{N-2}\times\{1\}$ respectively.
Again, by induction, these sets are contractible. A pair of contractible 
sets joined along contractible subsets is contractible.
\qed
\vspace{0.1in}

This should come as no surprise: in the limit as $N\to\infty$, the
reconfigurable system approximates the configuration space of a smooth
curve of fixed length which is positive in the sense that the 
curve is always nondecreasing in the horizontal and vertical 
components. That the (infinite dimensional) space of such smooth curves 
is contractible is easily demonstrated: given any such curve with 
endpoint fixed at the origin 
in the plane, pull the other end along the straight line connecting it
to the origin until the strand is taut. Then, rotate the line segment
rigidly until it is, say, vertical. This is a continuous deformation 
on the space of all smooth positive curves of fixed length to a 
single vertical segment. 
}\end{example}

It is certainly not the case that an arbitrary reconfigurable system 
possesses such convergence properties: the manner in which 
one refines the states is important. Still, we conjecture that 
state complexes can often be viewed as ``discretizations'' of some 
underlying smooth configuration space. 

\section{The geometry of $\scpx$}
\label{sec_Geometry}

\begin{figure}[hbt]
\begin{center}
\includegraphics[angle=0,width=3.2in]{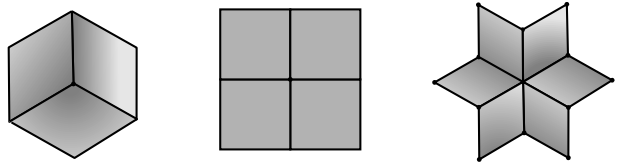}
\caption{Curvature about a vertex in a 2-d Euclidean cubical complex: 
positive, zero, and negative.}
\label{fig_Curvature}

\end{center}
\end{figure}

There are several natural ways to measure distances in state complexes.
We first discuss the geometry arising from considering each
cube of $\scpx$ to be Euclidean (i.e., flat), with unit side length;
we call $\scpx$ with this metric a {\em Euclidean cube complex}.
However, this does not imply that the complex, as a whole, is
flat. Indeed, non-zero curvature can be concentrated at
places where several cells meet. A simple example appears in 
Fig.~\ref{fig_Curvature}: here, a surface built from flat
2-cells can be seen to have curvature which depends on the number 
of 2-cells incident to a vertex. Four incident cells implies zero
curvature; three cells implies positive curvature; and five or
more cells implies negative curvature. For a two-dimensional complex, 
this is equivalent to computing the total angle about a vertex. 
%
%

Such an extension of curvature to general metric spaces is made 
precise in Gromov's work on curved metric spaces \cite{Gromov} (extending the 
classical work of Alexandrov, Busemann, and others) in which 
triangles with geodesic edges are used to measure curvature bounds.
In brief, let $X$ be a metric space and $p\in X$ a point. To 
bound the curvature of $X$ at $p$, consider a small triangle 
$T$ about $p$ with geodesic edges of length $a$, $b$, and $c$. 
Build a \df{comparison triangle} $T'$ in the Euclidean plane whose
sides also have length $a$, $b$, and $c$ respectively. Choose 
a geodesic chord of $T$ and measure its length $d$. In $T'$, 
measure the length $d'$ of the chord whose endpoints correspond 
to those of the chord in $X$.

\begin{definition}
\label{def_NPC}
A metric space $X$ is \df{nonpositively curved} (or \df{NPC}) 
if for every sufficiently small geodesic triangle $T$ and for every 
chord of $T$, it follows that $d\leq d'$. 
\end{definition}
%
\begin{figure}[hbt]
\begin{center}
\psfragscanon
\psfrag{a}[][]{\Large{$a$}}
\psfrag{b}[][]{\Large{$b$}}
\psfrag{c}[][]{\Large{$c$}}
\psfrag{d}[][]{\Large{$d$}}
\psfrag{A}[][]{\Large{$a$}}
\psfrag{B}[][]{\Large{$b$}}
\psfrag{C}[][]{\Large{$c$}}
\psfrag{D}[][]{\Large{$d'$}}
\psfrag{X}[][]{\Large{$X$}}
\psfrag{E}[][]{\Large{$\real^2$}}
\includegraphics[angle=0,width=3.2in]{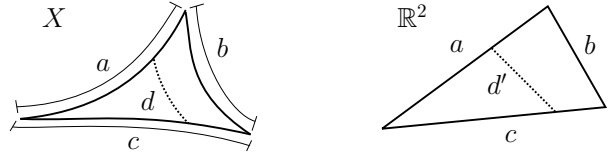}
\caption{Comparison triangles measure curvature bounds.}

\label{fig_CAT0}
\end{center}
\end{figure}

In other words, geodesic chords are no longer than Euclidean comparison 
chords. It should be stressed that the NPC property is very special and 
highly desirable. Indeed, being NPC implies a variety of topological 
consequences reminiscent of smooth nonpositively curved manifolds.

Despite the variety of (local) metamorphic systems, all state
complexes share this special geometric property. 
 
\begin{theorem}
\label{thm_NPC}
The state complex $\scpx$ of any local metamorphic system is
nonpositively curved.
\end{theorem}

The proof of this theorem is simple, but requires some additional
machinery. 

\begin{definition}{\em
\label{def_Link}
Let $X$ denote a complex (either simplicial or cubical) and let 
$v$ denote a vertex of $X$. The \df{link} of $v$, $\link{v}$, is
defined to be the abstract complex which has one $k$-dimensional 
simplex for each $(k+1)$-dimensional cube in $X$ incident to 
$v$. The boundary relations are those inherited from $X$: 
namely, the boundary of a $k$-simplex in $\link{v}$ represented
by a $(k+1)$-cube in $X$ is the set of all simplices 
represented by the faces of the $(k+1)$-cube. 
}\end{definition}

Links can be thought of as a simplicial version of the locus of points a
small fixed distance from the vertex $v$. 



\begin{definition}{\em 
\label{def_LinkCondition}
A Euclidean cube complex $X$ satisfies the \df{link condition} if, for each 
vertex $v\in X$, $\link{v}$ satisfies the following: for each $k$, if
any $k+1$ vertices in $\link{v}$ are pairwise connected by edges 
in $\link{v}$, then those vertices bound a unique $k$-simplex in $\link{v}$. 
}\end{definition}


An important and deep theorem of Gromov \cite{Gromov} asserts that a 
Euclidean cube complex is nonpositively curved if and only if 
it satisfies the link condition.  This criterion makes it easy to 
prove Theorem~\ref{thm_NPC}.
\vspace{0.1in}

\noindent{\sc proof of theorem~\ref{thm_NPC}:}~
Let $\state$ denote a vertex of $\scpx$. Consider the link
$\link{\state}$. The 0-cells of the $\link{\state}$ correspond
to all edges in $\scpx$ incident to $\state$; that is, 
actions of generators admissible at the state $\state$. A $k$-cell of 
$\link{\state}$ is thus a commuting set of $k+1$ of these actions
based at $\state$. The interpretation of the link condition for 
a state complex is as follows: if at $\state$ one has a 
set of $k+1$ admissible actions, 
$\{(\gen_{\alpha_i},\action_{\alpha_i})\}_{i=1}^{k+1}$, of which each pair 
commutes, then the full set of $k+1$ generators must commute 
[existence of the $k$-simplex in the link]. Furthermore, they
must commute in a unique manner [uniqueness of the $k$-simplex in the 
link]. 

The proof of existence follows directly from Definition~\ref{def_Commute}:
any collection of pairwise commutative actions is totally commutative.  
We therefore have a $(k+1)$-dimensional cell in $\scpx$ which is the 
equivalence class $[\state;(\action_{\alpha_i})_{i=1}^{k+1}]$.
This is the representative in $\scpx$ of the $k$-simplex in $\link{\state}$. 
To show uniqueness, consider any other $k$-simplex in $\link{\state}$ 
which has the same vertex set. This must correspond to 
a $(k+1)$-dimensional cube in $\scpx$ with actions 
$(\gen_{\alpha_i},\action_{\alpha_i})_{i=1}^{k+1}$ (up to some 
permutation) based at $\state$. From 
Definition~\ref{def_StateCpx}, this cell must be the same
equivalence class, namely $[\state;(\action_{\alpha_i})_{i=1}^{k+1}]$.
\qed\vspace{0.1in}

\begin{figure}[hbt]
\begin{center}
\includegraphics[angle=0,width=3.8in]{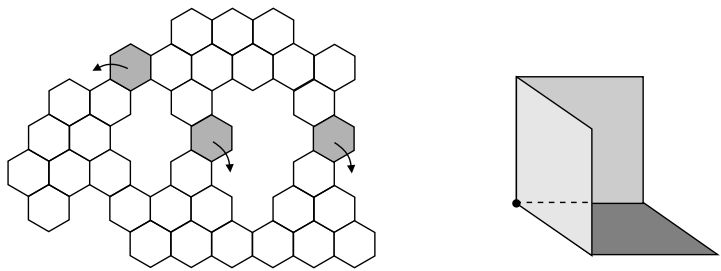}
\caption{A generator which allows for local disconnection [left] 
admits configurations [center] for which pairs of actions leave
the aggregate connected, but triples do not; this non-local system 
has positive curvature in the state complex at this state [right].}

\label{fig_Nonlocal}
\end{center}
\end{figure}

\begin{example}{\em
To see how the NPC property can fail, consider the following non-local 
metamorphic system. Recall the generator for the planar hexagonal
system presented in Fig.~\ref{fig_HexEx}[right] which (along with 
its rotations) allows for local disconnection of the aggregate. If 
we add to this system a global rule that requires the aggregate to be 
connected (for, say, considerations of power transmission), then we no 
longer have a local system, and positive curvature may exist. 
Fig.~\ref{fig_Nonlocal} shows a configuration in 
which three actions of local generators act to disconnect the 
aggregate locally but not globally. These actions commute pairwise, 
and any two do not disconnect globally. However, performing all three
actions disconnects the space and leads to an illegal state. Therefore, 
the corresponding state complex for this non-local system has a
``corner'' of positive curvature as a local factor. Note: the 
state complex will not be two-dimensional here, but will be locally a 
product of this with another cubical complex. The positive curvature
persists.
}\end{example}

\section{Geodesics and time on $\scpx$}
\label{sec_Time}

Besides displaying a unifying geometric feature, the nonpositive curvature
has implications for path-planning, and hence, the shape-planning problem. 

\begin{cor}
\label{cor_Unique}
Each homotopy class of paths connecting two given points of a state
complex contains a unique shortest path.
\end{cor}

\pf~
This is well-known for NPC spaces. The only difficulty lies in proving
that the $d\leq d'$ inequality for sufficiently small triangles 
implies the $d\leq d'$ inequality for {\em all} geodesic triangles 
which are contractible in the space: see \cite{BH}.  
Assume, then, that this inequality holds in general, and   
consider a pair of distinct homotopic shortest paths from 
points $p$ to $q$ in $\scpx$. Choose any point $r$ on one of the
two paths. The two halves of this path bisected by $r$ are themselves
shortest paths from $p$ to $r$ and $r$ to $q$ respectively. Thus, there is
a geodesic triangle in $\scpx$ whose comparison triangle in the 
Euclidean plane is a degenerate straight line from $p'$ to $q'$. 
The $d\leq d'$ inequality applied to the segment from $r$ to its
corresponding point on the other geodesic shows that these points
coincide; therefore the two shortest paths from $p$ to $q$ in 
$\scpx$ are identical.
\qed\vspace{0.1in}

\begin{figure}[hbt]
\begin{center}
\includegraphics[angle=0,width=3.3in]{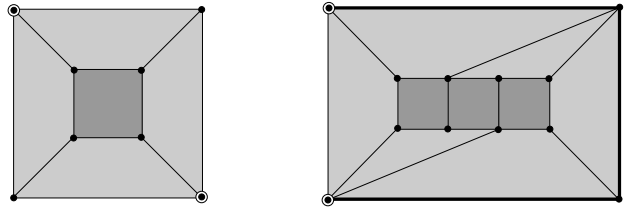}
\caption{[left] A 2-d cube-complex (with positive curvature) possessing 
distinct homotopic shortest paths between a pair of marked points --- note 
all edges have length one;  [right] 
a 2-d cube-complex (with positive curvature) possessing an edge path
(the thick line on the boundary) which is a locally (but not globally)
shortest path. Any cube path near this path is strictly longer.}
\label{fig_PC}
\end{center}
\end{figure}

Fig.~\ref{fig_PC}[left] gives a simple example of a 2-d cubical
complex with positive curvature for which the above corollary fails.

\begin{figure}[hbt]
\begin{center}
\includegraphics[angle=0,width=4in]{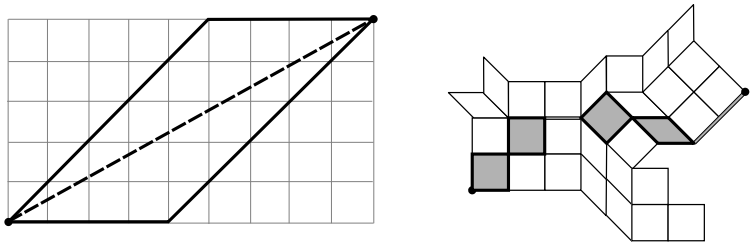}
\caption{[left] The true geodesic lies within an envelope of time-minimizing 
paths. No minimal edge-paths are time-minimizing; [right] a normal 
cube-path in an NPC complex (after \cite{NR}).}

\label{fig_Geods}
\end{center}
\end{figure}

This corollary is a key ingredient in the applications of NPC geometry
to path-planning on a configuration space, since 
one expects geodesics on $\scpx$ to coincide with optimal 
solutions to the shape planning problem. 
%
However, in the context of robotics applications, the goal of solving the 
shape-planning problem is {\em not} necessarily coincident with the 
geodesic problem on the state complex. Fig.~\ref{fig_Geods}[left] 
illustrates the matter concisely. Consider a portion of a state 
complex $\scpx$ which is planar and two-dimensional. To get from 
point $p$ to point $q$ in $\scpx$, any edge-path which is weakly
monotone increasing in the horizontal and vertical directions is 
of minimal length in the transition graph. The true geodesic is, of
course, the straight line, which is hardly a discrete object and 
thus more difficult to compute. 

Given the assumption that {\em each elementary move can be executed 
at a uniform maximum rate}, it is clear that the true geodesic 
on $\scpx$ is \df{time-minimal} in the sense that the elapsed time 
is minimal among all paths from $p$ to $q$. 
However, there is an envelope of non-geodesic paths which are yet 
time-minimizing.
Indeed, the true geodesic in Fig.~\ref{fig_Geods}[left] ``slows down''
some of the moves unnecessarily in order to maintain the constant slope. 
%
%

This leads us to define a second metric on $\scpx$, one which measures
elapsed time.
Namely, instead of the Euclidean metric on the cells of $\scpx$, consider
the space $\scpx$ with the $\ell^\infty$ norm on each cell.  (This is
also called the supremum norm:  a vector is measured by the maximum of 
its components in each coordinate direction.)  The
geodesics in this geometry represent reconfiguration paths which are 
{\em time minimizing}.  Using the results of \cite{NR}, one can prove 
that these geodesics are easily described using the notion of a 
\df{cube path}. 

\begin{definition}
\label{def_CubePath}
A \df{cube path} from a vertex $v_0$ to a vertex $v_N$ is an ordered 
sequence of closed cubes $\{C_i\}_1^N$ in $\scpx$ which satisfy 
(1) $C_{i}\cap C_{i+1}=v_i$ for some vertex $v_i$; 
and (2) $C_i$ is the smallest cell of $\scpx$ containing $v_{i-1}$
and $v_{i}$. A cube path is said to be \df{normal} if in addition 
(3) for all $i$, $\St(C_{i})\cap C_{i+1} =v_i$, where 
$\St(C_{i})$ is the \df{star} of 
$C_{i}$ (the union of all closed cubes, including $C_i$, which have 
$C_i$ as a face).
\end{definition}

Roughly speaking, a normal cube path is one which uses the 
highest dimensional cubes as early as possible in the path. 

\begin{theorem}
\label{thm_Geodesics}
In any state complex $\scpx$, there exists a unique normal cube path
from $p$ to $q$ in each homotopy class. The chain of diagonals 
through this cube path minimizes time among all paths in its homotopy 
class. 
\end{theorem}

This result follows from Theorem~\ref{thm_NPC} and a result 
of \cite{NR} (or, see the constructive algorithm of Section~\ref{sec_Alg}
below for a proof of existence). 

This result (or, alternatively, the constructive algorithm in the 
next section) also implies that, in the category of homotopic cube-paths,
there is no such thing as a strictly local minimum for length: if 
a cube path in $\scpx$ has the property that all nearby cube paths 
are longer (more cubes), then this cube path is a normal cube path
and indeed is the shortest cube path within the homotopy class of paths.
See Fig.~\ref{fig_PC}[right] for a simple example of a cube complex
with some positive curvature that has a cube path which is 
locally of minimal length among cube paths, but not globally so. 

This is very significant. By beginning with any path in the state 
complex, say, one obtained via a fast distributed algorithm for 
shape-planning, one can employ a gradient-descent curve shortening
on the level of cube paths. The above results imply that any 
algorithm which monotonically reduces length must converge to 
the shortest path (in that homotopy class) and cannot be hung up
on a locally minimal cube path. The presence of local minima is
a persistent problem in optimization schemes: nonpositive curvature
is a handy antidote.

\section{Optimizing paths}
\label{sec_Alg}

In cases where the state complex is sparse --- local principal cells
being of high dimension with few neighbors --- it is possible to 
compress the size of the transition graph significantly. However, 
shape-planning via constructing all of $\scpx$ and determining 
geodesics is, in general, computationally infeasible: the total 
size of the state complex is often exponential in the number of 
cells in the aggregate. We therefore assume that some shape trajectory 
has been determined (by a perhaps ad hoc or distributed method) 
and turn to the problem of optimizing
this trajectory. The nonpositive curvature of $\scpx$ allows for 
a time-optimization which does not require explicit construction of
$\scpx$. We detail an algorithm for transforming 
any given edge-path to a time-optimal normal cube path. 

From Definition~\ref{def_StateCpx}, an $n$-dimensional cube $C$ of 
$\scpx$ can be represented by a set of $n$ commutative actions
$\{\action_i\}$ along with an admissible state $\state$. In the following 
algorithm, we suppress the state for notational convenience and 
consider $C$ as the set of actions. Addition and subtraction is 
defined by adding or taking away admissible commutative actions to 
the list. 

\begin{alg}[{\tt TimeGeodesic}]
\label{alg_Time}
{\sf
Given: a cube path $\CC=\{C_i\}_{i=1..N}$ in $\scpx$.

\begin{tabular}{rl}
1: &  Let $N:=$ {\tt Length}$(\CC)$. \\
2: &  Call {\tt ShrinkCubePath}$(\CC)$. \\
3: &  If {\tt Length}$(\CC)<N$ then 1: else {\tt stop}.
\end{tabular}
}\end{alg}

\begin{alg}[{\tt ShrinkCubePath}]
\label{alg_Short}
{\sf
Given: a cube path $\CC=\{C_i\}_{i=1..N}$ in $\scpx$.

\begin{tabular}{rl}
1: &  Let $i=1$. \\
2: &  Let $X:=$ {\tt Commute}$(C_{i};C_{i+1})$. \\
3: &  Update $C_{i}:=C_{i}+X$; \\
4: &  Update $C_{i+1}:=C_{i+1}-X$. \\
5: &  Call {\tt ExciseTrivial}$(C_{i+1})$. \\
6: &  Call {\tt CommonEdge}$(C_{i-1},C_i)$. \\
7: &  Call {\tt ExciseTrivial}$(C_{i-1};C_{i})$. \\
8: &  If $X=\emptyset$ then $i=i+1$. \\
9: &  If $C_{i+1}\neq\emptyset$ then 2: else {\tt stop}.\\
\end{tabular}
}\end{alg}

\begin{sub}[{\tt Commute}]
{\sf
Given: a pair of cubes $C_j=\{\action^j_{\alpha_i}\}_{i=1..m}$ and 
$C_{j+1}=\{\action^{j+1}_{\beta_i}\}_{i=1..n}$,

\begin{tabular}{rl}
1: & Let $S:=\bigcup_{i}\support(\action_{\alpha_i}^j)$    \\
2: & Let $T:=\bigcup_{i}\trace(\action_{\alpha_i}^j)$   \\
3: & For $i=1..n$ return $\action^{j+1}_{\beta_i}$ if  \\
 & 3.1: $S\cap\trace(\action^{j+1}_{\beta_i})=\emptyset$ ; and \\
 & 3.2: $T\cap\support(\action^{j+1}_{\beta_i})=\emptyset$. 
\end{tabular}
}\end{sub}



Subroutine {\tt ExciseTrivial} checks a cube 
for an empty list, removes these from the path, and reindexes the cube
path, thus reducing the length. 
Subroutine {\tt CommonEdge} checks a cube against adjacent cubes in 
the sequence for common edges and deletes these, returning a genuine
cube path (recall Definition~\ref{def_CubePath}).

Subroutine {\tt Commute} takes as its argument a 
pair of cubes $C_i$ and $C_{i+1}$ and returns those elements of 
$C_{i+1}$ which commute with {\em all} elements of $C_i$. 
The following lemma is a manipulation of the definitions. 
\begin{lemma}
\label{lem_Star=Comm}
The result of {\tt Commute}$(C_i;C_{i+1})$ is precisely the set of 
edges in $\St(C_i)\cap C_{i+1}$.  
\end{lemma}


%
\begin{figure}[hbt]
\begin{center}
\psfragscanon
\psfrag{1}[][]{\Large{$\CC$}}
\psfrag{2}[][]{\Large{$1$}}
\psfrag{3}[][]{\Large{$2$}}
\psfrag{4}[][]{\Large{$3$}}
\includegraphics[angle=0,width=3.5in]{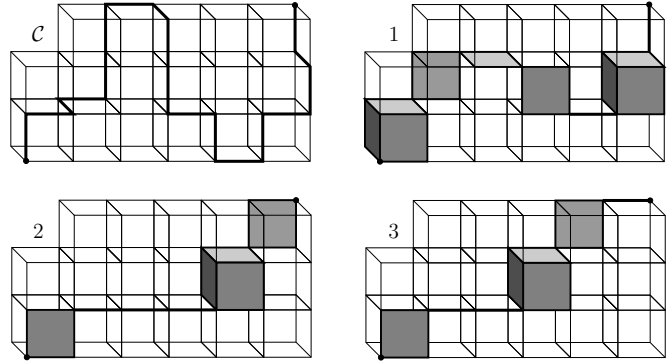}
\caption{Three rounds of shortening an edge path $\CC$. Cube path 
length stabilizes after two calls to {\tt ShrinkCubePath}. 
Three more calls produces a normal cube path.}

\label{fig_Algorithm}
\end{center}
\end{figure}
%

\begin{theorem}
Given a cube path $\CC$, Algorithm~\ref{alg_Time} computes a 
globally time-optimal path in the homotopy class of $\CC$. 
\end{theorem}

\pf~
Recall that in this setting, the length of a cube path refers
to the number of cubes in the path; this equals the time 
required to execute the reconfiguration.

Repeated calls to Algorithm {\tt ShrinkCubePath} eventually  
leave a cube path fixed. To show this, note that the 
integer-valued function $f(\CC):=\sum_ii\cdot\dim(C_i)$ decreases 
in each call to {\tt ShrinkCubePath} which changes the path. 
Lemma~\ref{lem_Star=Comm}
implies that {\tt ShrinkCubePath} leaves a cube path fixed if and only 
if it is a normal cube path. From Theorem~\ref{thm_Geodesics}, this 
yields a time-optimal path. 

\begin{figure}[hbt]
\begin{center}
\psfragscanon
\psfrag{k}[][]{\Large{$C_{i+1}$}}
\psfrag{i}[][]{\Large{$C_{i-1}$}}
\psfrag{j}[][]{\Large{$C_{i}$}}
\psfrag{E}[][]{\Large{$=$}}
\includegraphics[angle=0,width=2.6in]{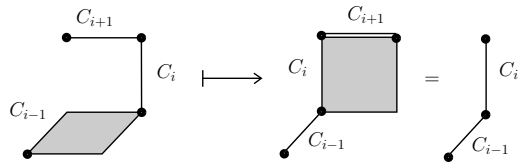}
\caption{Schematic of exchanging actions in sequential rounds.
Each dimension in this cartoon represents a multi-dimensional cube.}
\label{fig_Exchange}
\end{center}
\end{figure}

However, Algorithm {\tt TimeGeodesic} calls {\tt ShrinkCubePath} only
until the {\em length} of the cube path is unchanged.  Thus it 
remains to show that once the length of a cube path 
in unchanged by a call to {\tt ShrinkCubePath}, this is equal 
to the length of the associated normal cube path. To show this, 
assume that {\tt ShrinkCubePath} stops without
excising any trivial cubes (i.e., shortening the length). 
We show that further calls merely redistribute actions earlier in 
the list $\CC$ without deleting any permanently. 
Without a loss of generality, assume that the first call to 
{\tt ShrinkCubePath} does not change
the length of the cube path, but that in the subsequent call, the cube 
$C_{i+1}$ is eliminated via $C_i$ commuting with a portion of
$C_{i-1}$: see the schematic of Fig.~\ref{fig_Exchange}.

Since $C_{i+1}$ is not eliminated in the first call, there must have been 
actions in $C_i$ which did not commute with those of $C_{i-1}$ but
which are subsequently pushed backward in the second call. In addition, 
all the actions of $C_{i+1}$ are contained in (``parallel to'' in the figure) 
the subset of $C_{i-1}$ which commutes with $C_i$ in the second call. 
This being the case, the commutativity of $C_{i+1}$ with $C_i$ would 
have occurred in the first call: contradiction.
\qed\vspace{0.1in}

It is not surprising that Algorithm~\ref{alg_Time} 
converges to a {\em locally} time-minimal solution. The interesting 
implication is that the cube-path obtained 
is the {\em global} minimum for all paths obtainable from the initial:
there is no way to get a quicker reconfiguration path by first
lengthening the path and then shortening.  This is the boon
of nonpositive curvature.  Fig.~\ref{fig_Algorithm}
gives an example of paths generated by Algorithm~\ref{alg_Time}. 

%
The complexity of Algorithm~\ref{alg_Time} depends on both the
catalogue of generators for the system and the length $N$ of the 
input edge path.  However, the dependence on the catalogue arises 
only in Subroutine {\tt Commute}, which tests various pairs of subsets
of the workspace for disjointness; the sizes of these subsets are
governed by the generators.  Assuming a fixed catalogue, then, 
there is a constant bound on the time required for a single call 
to {\tt Commute}, and we may analyze the complexity of 
Algorithm~\ref{alg_Time} as a function of $N$ alone.  

Notice that Algorithm~\ref{alg_Time} calls 
{\tt ShrinkCubePath} at most $N$ times, since it stops as soon
as the length of the cube path fails to decrease.  
%
Each time an individual loop of {\tt ShrinkCubePath} is executed, the
quantity $\sum
\dim(C_j)$ decreases.  Since this quantity
equals $N$ initially, {\tt ShrinkCubePath} calls the subroutines
{\tt Commute}, {\tt ExciseTrivial}, and {\tt CommonEdge} each at
most $N$ times.  Each of these has a constant running time, with
{\tt Commute} being the only one depending on the particulars of
the metamorphic system.  Therefore the complexity of the entire 
Algorithm~\ref{alg_Time} is $O(N^2)$, with the constant determined 
by the running time of {\tt Commute}.

The worst case for Algorithm~\ref{alg_Time} seems to be achieved in
a totally flat 2-d state complex by a path consisting of two perpendicular
segments of length $N/2$.  (This case requires $N^2/2$ runs of 
{\tt ShrinkCubePath}.)  The presence of true negative curvature
in a state complex greatly increases the speed of convergence. 
On the other hand, a flat state complex is desirable for other reasons.

It remains an important computational question to determine 
whether the homotopy class of the initial path (given, say, from
a distributed algorithm) is optimal in the sense that its
geodesic is the shortest among all homotopy classes. 
Interestingly, in contrast to the previous paragraph, this
problem becomes quite a bit more difficult in the presence of 
negative curvature.  From this point of view it is preferable
to have a flat state complex.


\section{Discussion}

Our principal contributions are:
\begin{enumerate}
\item
{\em A mathematical definition} of a metamorphic system which 
encompass many models currently studied and suggests 
seemingly unrelated (and often simpler) systems. 
Given the difficulty of building large metamorphic
systems, simpler examples possessing the same formal structure 
may be valuable. 
\item {\em The state complex}, whose
naturality is manifested on the level of its 
topology ($\scpx$ can be homotopically simple) and its geometry 
(non-positive curvature is universal, mathematically helpful,
and highly desirable for shape planning).
\end{enumerate}

There are drawbacks to this approach. Primary among them are the dual 
dilemmas that shape planning is inherently complex, and that there 
are many types of reconfiguration possible. A state complex 
approach is not meant for all systems. Indeed, 
it is possible to design degenerate metamorphic systems with 
little to no commutativity. Nevertheless, paying attention to the geometry
lurking behind our higher-dimensional versions of transition graphs 
leads to a nontrivial result on the optimality of path-shortening. 

As an interesting side-note, we observe that in the present context,
the optimal reconfiguration path is not always coincident with a
geodesic on the configuration space. This principle holds
in more generality.

Finally, in this initial work, we have focused only on the first-order
problem of shape planning. We hope that the mathematical framework 
here suggested finds uses in more sophisticated task-planning problems 
for metamorphic and reconfigurable robots. 

\appendix 

\section{Shape complexes}                         

State complexes, despite their relative simplicity over the transition
graph, nevertheless typically contain a large number of cells.  For
lattice-based systems, and in some other cases as well, a much smaller 
complex called the \emph{shape complex} can be
used to encode all of the local reconfigurations, 
without losing any essential information from the state complex. 
The idea is to exploit the symmetries of the domain.

Consider the 
(lattice-based) pivoting hex system of Example~\ref{ex_Hex}.
The underlying lattice has translational symmetry in two independent
directions; so for instance, in the idealized case where the workspace 
is the entire (infinite) lattice with no obstacles, the state complex 
$\scpx$ shares these symmetries.  If we take a quotient of $\scpx$ by 
the actions of these two translations, we obtain the shape complex
$\shcpx$.  (The precise definition follows shortly.)

Observe that $\shcpx$ is much smaller than $\scpx$; indeed even if the
workspace is infinite, the complex $\shcpx$ is compact (provided
the system consists of connected sets of, say, $N$ cells).
Yet, $\shcpx$ carries substantially the same information as $\scpx$.
Whereas $\scpx$ keeps track of both the shape and the location of
the aggregate, the quotient $\shcpx$ ignores the location.
 
Note that no information about the system is lost in the
process of forming the quotient; $\scpx$ can be completely
reconstructed from $\shcpx$.  Indeed, $\scpx$ is a
\emph{covering space} of $\shcpx$ (with covering
transformation group equal to the group of symmetries of
the lattice).  In particular $\shcpx$ and $\scpx$ share
the same universal covering space, so $\shcpx$ (just like
$\scpx$) is non-positively curved.

Here is the definition of the shape complex; compare 
with Definition~\ref{def_StateCpx}.

\begin{definition}{\em
\label{def_ShapeCpx}
The \df{shape complex} $\shcpx$ of a (lattice-based) local 
reconfigurable system is the abstract cube complex 
whose $k$-dimensional cubes are equivalence classes
$[\state;(\Phi_{\alpha_i})_{i=1}^k]$ where
\begin{enumerate}
\item
        $(\Phi_{\alpha_i})_{i=1}^k$ is a
        $k$-tuple of commuting actions of generators $\phi_{\alpha_i}$;
\item
        $\state$ is some state for which all the actions
        $(\Phi_{\alpha_i})_{i=1}^k$ are admissible; and
\item
$         [\state_0;(\Phi_{\alpha_i})_{i=1}^k]
        =[\state_1;(\Phi_{\beta_i})_{i=1}^k]$
        if and only if there is a translation $T:\L\to\L$ such that 
	the list $(\Phi_{\beta_i}\circ T)$ is a permutation of 
	$(\Phi_{\alpha_i})$, and 
	$\state_0 =\state_1 \circ T$ on the set
$        \workspace - \bigcup_i\Phi_{\alpha_i}(\support(\gen_{\alpha_i}))$ .
\end{enumerate}

The boundary of a $k$-cube is obtained just as it is in $\scpx$.
}\end{definition}

\begin{example}{\em
\label{ex_ShapeCpx}
The shape complex for the system of Example~\ref{ex_Hex} (with the 
generator of Fig.~\ref{fig_HexEx}[left]) having a total of $N=3$
occupied cells is shown in Figure~\ref{fig_ShapeCpx}.  There are
nine square cells.  Two pairs of edges are identified according to 
the matching arrows, yielding a Mobius strip; additionally, the 
three black vertices are identified to one, as are the three white
vertices.  The shape complex is therefore a Mobius strip with some
of the boundary points identified.
\begin{figure}[hbt]
\begin{center}
\includegraphics[angle=0,width=1.35in]{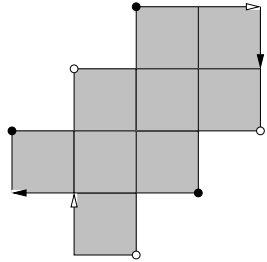}
\caption{The shape complex for a pivoting hex system consisting of
3 robotic cells.}
\label{fig_ShapeCpx}
\end{center}
\end{figure}
}
\end{example}
 
%

\indent\textbf{Macros.}
The most obvious advantage of the shape complex is that the
discovery of a single shortest path in $\shcpx$ simultaneously 
solves several state-changing problems in the complex $\scpx$.
This is particularly helpful in systems which require frequent
transportation within the workspace (perhaps to perform 
various tasks in different areas). 
For instance, suppose the workspace includes several 
narrow corridors through which the robot must travel at various 
times.  At each encounter with such a corridor, it must change 
to a shape which is skinny enough to navigate the passageway.  
But this is the same problem at each corridor, even though it 
occurs in different places in the state complex:  the three hexagons
in Example~\ref{ex_ShapeCpx} can move from a triangular shape
to a linear shape by the same sequence of moves, wherever the problem 
may arise in $\L$.  

In the shape complex, we view this as a single shortest-path 
problem, whose solution we may \emph{lift} to various locations 
in the domain.  In this sense a shortest path $P$ in $\shcpx$ can 
be viewed as a macro for solving state-changing problems in $\scpx$.  

A drawback to the shape complex is that it neglects the shape
of the workspace $\workspace$.  In the presence of obstacles or at the
edge of $\workspace$, a path $P$ in $\shcpx$ may fail to lift to
a path in $\scpx$.  Thus before lifting a path $P$ one must
check that the actions are admissible at the appropriate locations
in $\workspace$.


\bibliographystyle{plain}

\end{document}